\documentclass{article}

\usepackage{arxiv}

\usepackage[utf8]{inputenc} 
\usepackage[T1]{fontenc}    
\usepackage{hyperref}       
\usepackage{url}            
\usepackage{booktabs}       
\usepackage{amsfonts}       
\usepackage{nicefrac}       
\usepackage{microtype}      
\usepackage{lipsum}
\usepackage{graphicx}
\usepackage[numbers]{natbib}
\usepackage{amsmath}
\usepackage{algorithm}
\usepackage{algpseudocode}
\graphicspath{ {./images/} }

\title{Optimal Power Allocation and Sub-optimal Channel Assignment for Downlink NOMA System Using Deep Reinforcement Learning}

\author{
  WooSeok Kim \\
  Department of Computer Science \\
  Sangmyung University \\
  \texttt{3suksw@gmail.com} \\
  \And
  Jeonghoon Lee\\
  Department of Game Design and Development \\
  Sangmyung University \\
  \texttt{2jh0926@naver.com} \\
  \And
  Sangho Kim \\
  Department of Computer Science \\
  Sangmyung University \\
  \texttt{ghtkdrla321@naver.com} \\
  \And
  Taesun An\\
  Department of Computer Science \\
  Sangmyung University \\
  \texttt{asdgqe1@gmail.com} \\
  \And
  WonMin Lee\\
  Department of Computer Science \\
  Sangmyung University \\
  \texttt{wonmin98@naver.com} \\
  \And
  Dowon Kim \\
  Department of Computer Science \\
  Sangmyung University \\
  \texttt{wlsgur479@gmail.com} \\
  \And
  Kyungseop Shin\\
  Department of Computer Science \\
  Sangmyung University \\
  \texttt{ksshin@smu.ac.kr} \\
}

\begin{document}
\maketitle
\begin{abstract}
In recent years, Non-Orthogonal Multiple Access (NOMA) system has emerged as a promising candidate for multiple access frameworks due to the evolution of deep machine learning, trying to incorporate deep machine learning into the NOMA system.
The main motivation for such active studies is the growing need to optimize the utilization of network resources as the expansion of the internet of things (IoT) caused a scarcity of network resources.
The NOMA addresses this need by power multiplexing, allowing multiple users to access the network simultaneously.
Nevertheless, the NOMA system has few limitations.
Several works have proposed to mitigate this, including the optimization of power allocation known as joint resource allocation(JRA) method, and integration of the JRA method and deep reinforcement learning (JRA-DRL).
Despite this, the channel assignment problem remains unclear and requires further investigation.
In this paper, we propose a deep reinforcement learning framework incorporating replay memory with an on-policy algorithm, allocating network resources in a NOMA system to generalize the learning.
Also, we provide extensive simulations to evaluate the effects of varying the learning rate, batch size, type of model, and the number of features in the state.
\end{abstract}


\section{Introduction}
Over the past few years, rapid development in Internet of Things (IoT) has resulted in a drastical increase in network demands,
leading to the new challenge of guaranteeing massive connectivity and quality of service (QoS).
To fulfill such demands and challenges, recents studies are focusing on integrating artificial intelligence (AI) to networking system \cite{DLandNetworkingSurvey} \cite{smu}.
For instance,
\citet{DRL-hetero} used an AI to learn the optimal wireless resource allocation method for MAC protocols and
\citet{sleep/wake-up} focused on packet delay and power efficiency.

To be specific, there have been multiple attempts to use AI to fully use the advantages of Non-Orthogonal Multiple Access (NOMA).
Compared to a conventional technique called Orthogonal Multiple Access (OMA) which is to allocate network resources orthogonally, 
NOMA mainly utilizes the ability of successive interference cancellation (SIC) which enables differentiation of users through different power assignments even in the same resource block.
SIC is a technique which decodes received signals sequentially, treating unrelated signals as interference and then remove the signals~\cite{JRA_DRL}.
NOMA is a spectrum-efficient wireless networking technique, allowing multiple users to share common resources such that time and frequency
and it is anticipated that NOMA will play a pivotal role in 5G era and future wireless networking system as the technological advance in AI.

Although, NOMA has some limitations in IoT environments.
Representatively, solution of assigning channels and allocating powers is known to be NP-hard~\cite{np-hard} and the complexity of the system increases as the nature of dynamic environment and the SIC.
Reinforcement learning is suggested as a potential solution to resolve such issues.
Not only reinforcement learning has the ability to process complex system, but also it can learn the optimal policy off of dynamic environmental systems, allocating channels and assigning powers.
Furthermore, researchers propose various algorithms to enhance the performance for optimal resource allocation problems.
Solving a power allocation and an assigning channel problems is the key to the optimal resource allocation in NOMA system.
\citet{JRA_DRL} suggested a power assignment method improving channel gain, and also proposed a joint resource allocation (JRA) and channel allocation method to maximize the NOMA system using DRL framework.

\citet{uplinkNOMA-RL} utilized the potential of the NOMA power domain.
The following paper proposed an efficient and optimized algorithm to enhance IoT connectivity , utilizing DRL and State-Action-Reward-State-Action (SARSA).
SARSA is an on-policy algorithm in which the agent selects an action based on the current policy then evaluates and update the policy based on the action taken.
The paper shows that the IoT networking utilizing NOMA outperforms the IoT networking using OMA system in terms of the number of processes agent can take in.

Under downlink NOMA system which is to estimate imperfect channels,
\citet{downlinkNOMA-power-imperfect} proposed an approach to allocate power.
Considering the two information, channel estimation and Mean Squared Error (MSE), the objective is to set the upper bound of System Outage Probability (SOP) using two users' throughput requirements.
Afterwards, in order to make the SOP minimized under overall power constraints, outage power allocation solution for two users is driven.
The solution can be driven within few calculations for power allocation coefficient, by the MSE of the channel estimation, and is way less complex than the previous outage or repetitive solutions.
The simulation results are showing that the proposed method achieves great performance in various transmission rate requirements resulting a SOP.

\citet{PA-mmWave} proposed a power allocation algorithm for one BS and two user clusters assigned to mmWave-NOMA system.
To be specific, the proposed algorithm meets the individual service quality constraint requirements and maximizes achievable sum rate (ASR) and energy efficiency (EE), in consequence, it formulated the optimization problems.
In order to guarantee stability of SIC, the algorithm added power order constraint which is commonly dismissed from the previous related works.
The algorithm divides the formulated problem into subproblems as clustering problem to make the problem easier to solve, deriving the solutions for ASR Maximization-based Power Allocation (ASRMax-PA) algorithm and EE Maximization-based PA (EEMax-PA).
The proposed ASRMax-PA (or EEMax-PA) algorithm outperforms than the latest methods in the aspect of ASR (or EE) and performs great in EE (or ASR) as well.
Not only that, the two proposed methods can assure the stability of SIC which is a critical factor for performance of NOMA system.

Although aforementioned researches do not explicitly point out the limitations that they have,
the simulations took place were conducted in a specific, restricted environment rather than a dynamic environment.
To address this issue, in this paper, we propose an effective framework where a Deep Reinforcement Learning (DRL) agent efficiently allocates limited networking resources in a downlink NOMA system.
The main difference between the proposed framework and previous works is that it uses an experience replay memory to generalize learning, rather than solving the on-policy problem with traditional policy gradient methods.
Policy gradient methods evaluate and update the policy every iteration and
experience replay is to save series of experiences that agent had and sample the experiences with batch resulting a generalized learning.
See Section~\ref{sec:rl} for more detailed explanation why experience replay memory has been applied to this framework.
The goal of the agent is to learn a policy for a downlink NOMA system under various profiles to enhance understanding of the generalized NOMA system attempting for a maximum sum throughput (\textit{i.e.}, sum rate).

The verification of our proposed framework is conducted through multiple simulations with varying controls to assess its performance.
In detail, changes in the types of neural networks such that fully connected neural network (FCNN), convolutional neural network (CNN), and attention-based neural network (ANN), batch sizes, learning rates, and number of NOMA users are made.
Also, the comparisons between the frameworks such that Joint Resource Allocation (JRA), JRA-DRL, the proposed framework, and Exhaustive Search (ES) are conducted as well.
The paper will thoroughly analyze the experimental results by inspecting loss, loss convergence speed and resulted sum rate.
Especially, since the environment of networking is dynamic and ever-changing, the convergence speed is crucial in networking system.
The problem addressed in this paper is maximizing data throughput through efficient resource allocation.
Therefore, overall environmental definition will be settled first.

\textbf{Contributions}.
Since the JRA-DRL method learns the policy of the NOMA system directly from current experience, it may lack generality and be unable to handle various scenarios.
To address this, we have incorporated a policy gradient method and replay memory to enhance the generality.
Therefore, we have incorporated a policy gradient method and replay memory to enhance generality.
Additionally, deep learning techniques require extensive fine-tuning, such as changing the model architecture, tuning hyperparameters, and, in DRL, the design of the state directly influences the training results.
To demonstrate the effectiveness of our method, we provide extensive simulation results.
We make the following contributions:

\begin{itemize}
  \item Incorporation of policy gradient method and replay memory: 
    This approach avoids biased training since the experiences used for training are sampled from a replay memory, unlike the conventional policy gradient method, which trains on current experience and may result in overfitting.
    The use of replay memory results in more balanced and generalized learning of the NOMA system, reducing the variability in training outcomes while improving the convergent stability and reliability.
  \item Extensive simulation results:
    We observed that small changes in hyperparameters, model architectures, or the design of the DRL state lead to significant differences in simulation results.
    Therefore, we carefully evaluated a range of settings to identify the most robust configuration and provided numerous simulation results.
    This thorough analysis examines the impact of the DRL state size, which gradually increases the number of key features in the given NOMA system, convergence speed when modifying model architectures, and the fine-tuning of hyperparameters.
\end{itemize}

\section{System Model}
In this paper, we assume a downlink NOMA system where BS sends data to multiple users in wireless channels.
Given this environment, a comprehensive definition of overall environment is necessary.

In a wireless channel, note that power and channel information is required in order to describe the relationships between BS and users.
The BS's job is to distribute limited joint resources (\textit{i.e.}, channel and power) and multiplex signals, then transmit the multiplexed signals to users.
After users receive multiple signals which contain independent signals, users use decoder to perform SIC and specify the signal for their own.
Additionally, there is an upper bound for power allocation, as well as for channel bandwidth.
Given these constraints, by using two methods, JRA and DRL, an optimal solution allocating joint resources can be found.

\begin{figure*}
  \centering
  \includegraphics[width=1.0\linewidth]{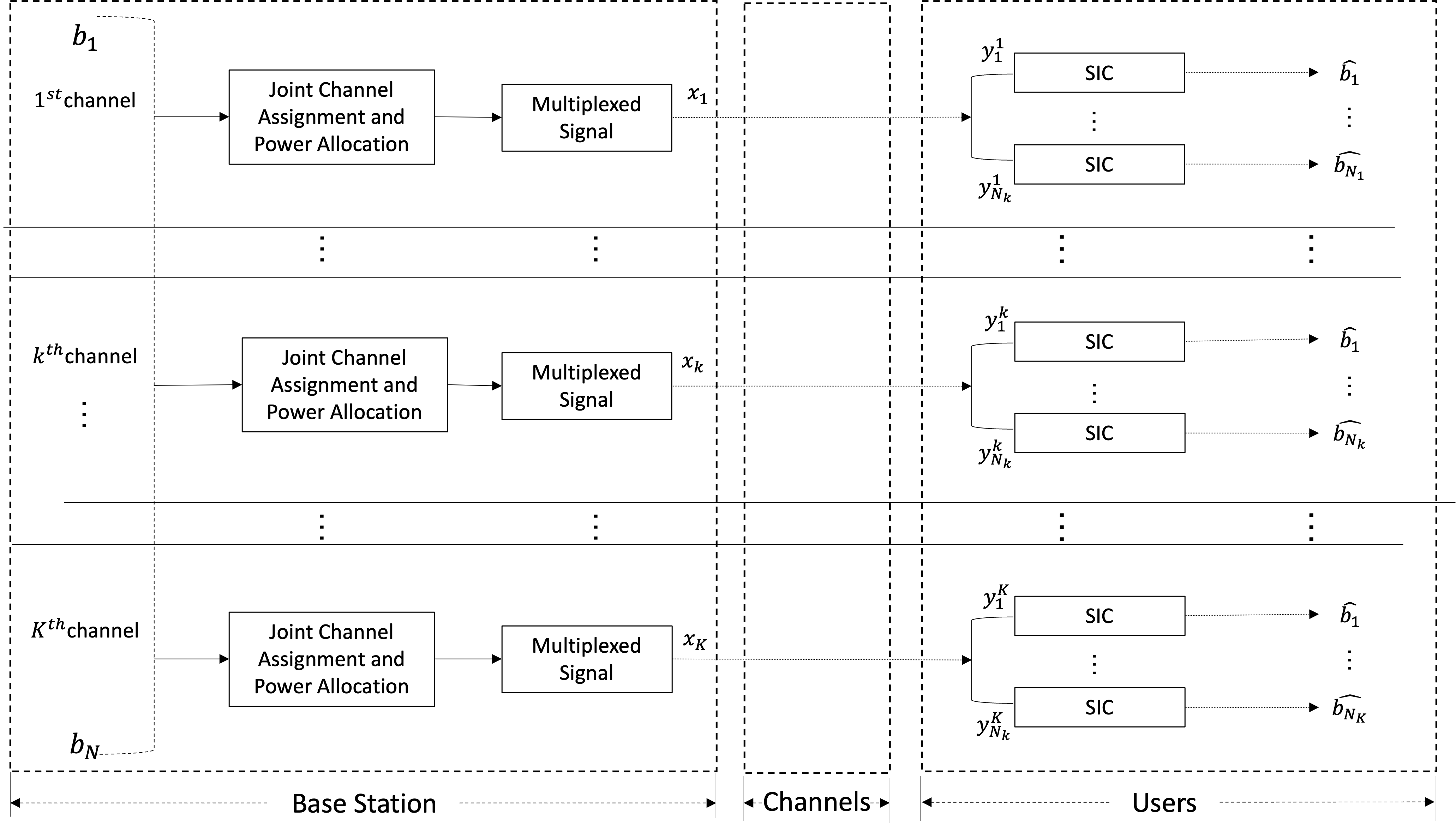}
  \caption{Block diagram illustrating the transmission of BS and reception of users of the downlink NOMA system.}
  \label{fig:1}
\end{figure*}

Figure.~\ref{fig:1} briefly illustrates the transmission and the reception between BS and users in a downlink NOMA system,
where we assume there are $N$ users and $K$ channels.
The total bandwidth is $B_{tot}$ and since all channels have the same bandwidth, the bandwidth for each channel is represented as $B_c = B_{tot} / K$.
Also, the number of users assigned to channel $k$ is $N_k$.
Consider a user $i$'s signal as $b_i$, then the BS will transmit multiple signals as follows,

\begin{equation} \label{eq-1}
  x_k = \sum_{i=1}^{N_k}{\sqrt{p_i^k}b_i}
\end{equation}

\noindent where $p_i^k$ is a power of $i$-th user assigned to $k$-th channel,
$x_k$ is a multiplexed signal from $k$-th channel.
As receiver receives signal from transmitter, the signal is corrupted by environmental noises.
The receiver will eventually receive a signal $y_k$ and can be written as,

\begin{equation} \label{eq-2}
  y_n^k =
  \sqrt{p_n^k} h_n^k b_n
  + \sum_{i=1, i \neq n}^{N_k}{\sqrt{p_n^k} h_n^k b_i}
  + z_n^k
\end{equation}

\noindent where $h_n^k$ is a $k$-th channel response between BS and $n$-th user, $z_n^k$ is user $n$'s Additive White Gaussian Noise (AWGN) with zero mean and variance of $\sigma^2_{z_k}$.
When the signals from multiple users are multiplexed as shown in (\ref{eq-1}) and the receiver's noise-added final received signal is given by (\ref{eq-2}),
SIC is applied to decode each users' signal, differentiating multiple signals.

In order for SIC to successfully be performed, channel-to-noise-ratio (CNR) should be considered which can be represented as $\Gamma_n^k = |h_n^k|^2 / \sigma_n^k$.
Since the greater power is assigned to users with lower CNR, according to the NOMA protocol, powers and CNRs on $k$-th channel can be ordered as such:

\begin{align*}
  \Gamma_1^k &> \Gamma_2^k > \ldots > \Gamma_n^k > \ldots > \Gamma_{N_k}^k, \\
  p_1^k &< p_2^k < \ldots < p_n^k < \ldots < p_{N_k}^k.
\end{align*}

This behavior allows other weak signals to be treated as noises while the signal with greater power to be decoded primarily.
Also the corresponding data rate is as follows,

\begin{equation}
  R_n^k(\Gamma_n^k, p_1^k, \ldots, p_n^k)
  = B_c \log_2 
  \left( 1 +
    \frac{p_n^k \Gamma_n^k}{1 + \sum_{i=1}^{n-1}{p_i^k \Gamma_n^k}}
  \right).
  \label{eq-3}
\end{equation}

As stated by
\citet{n_k_1} and \citet{n_k_2}, the number of users allocated in channel $k$ is fixed to two (\textit{i.e.}, $N_k=2$),
because the increase in $N_k$ directly affects the hardward implementation complexity and processing time.
Reflecting the mentioned change, the data rates for two users allocated at channel $k$ are represented as,

\begin{equation} \label{eq-4}
  \begin{aligned}
    R^k_1(\Gamma^k_1, p^k_1, p^k_2) &= B_c\log_2{(1+p^k_1\Gamma^k_1)}, \\
    R^k_2(\Gamma^k_2, p^k_1, p^k_2) &= B_c\log_2{(1+ \frac{p^k_2\Gamma^k_2}{1+p^k_1\Gamma^k_2})}.
  \end{aligned}
\end{equation}

Note that the problem we are trying to solve is to maximize the data rates of NOMA users. 
In other words, in this paper, we focus on maximizing sum rate (MSR) metric.
To maximize the sum rate, it is necessary to consider the data rates of all users.
This often means that it may be inevitable to sacrifice the data rates of few users in order to achieve overall higher sum rate.

As mentioned above, we assume two users can be allocated to each channel (\textit{i.e.}, $N$ = $2K$), and the objective is to maximize the sum throughput of users.
We assumed that there is a limit of total power $P_T$ for BS which needs to be distributed to all users across the channels.
It is important to note that the sum of users' power $p^k_1$ and $p^k_2$ must not exceed $P_T$ as follows,

\begin{equation}
  \sum_{k=1}^{K}{(p_1^k + p_2^k)} \leq P_T
\end{equation}

For power assignment, the JRA method~\cite{JRA} will be used which is an optimal power assignment solution using mathametical derivation.
Since JRA method is able to find the optimal power given the channel allocations and resources, we propose to apply JRA method alongside with DRL channel allocations.
The problem for MSR metric formulation can be written as,

\begin{equation}
  \begin{aligned}
    \max_{p_1, p_2} &\sum_{k=1}^{K}
    \left[
    R^k_1(p^k_1, p^k_2) + R^k_2(p^k_1, p^k_2)
    \right], \\
    s.t. \; & R_n^k \geq (R_n^k)_{min}, n = 1, 2, \forall k = 1, \ldots, K, \\
    & \sum_{k=1}^{K}(p_1^k + p_2^k) \leq P_T, \\
    & 0 \leq p_1^k \leq p_2^k, \forall k = 1, \ldots, K,
  \end{aligned}
  \label{eq-6}
\end{equation}

\noindent where $(R_n^k)_{min}$ is a minimum data rate requirement for user $n$ allocated to $k$-th channel.
In order to solve the (\ref{eq-6}), the optimization problem decomposes into the following subproblems for each channel $k$,

\begin{equation}
  \begin{aligned}
    \max_{p^k_1, p^k_2} \; & R^k_1(p^k_1, p^k_2) + R^k_2(p^k_1, p^k_2) \\
    s.t. \; & R_n^k \geq (R_n^k)_{min}, n = 1, 2, \forall k = 1, \ldots, K, \\
            & p^k_1 + p^k_2 = q^k, \\
    & 0 \leq p_1^k \leq p_2^k, \forall k = 1, \ldots, K,
  \end{aligned}
  \label{eq-7}
\end{equation}

\noindent where $q^k$ is a power budget for channel $k$.

As
\citet{JRA} proposed, the solution for MSR metric is given by solving the subproblems,

\begin{equation}
  \begin{aligned}
    p_1^k &= \frac{\Gamma_2^k q^k - A_n^k + 1}{A_2^k \Gamma_2^k}, \\
    p_2^k &= q^k - p_1^k,
  \end{aligned}
  \label{eq-8}
\end{equation}

\noindent where $A_n^k = 2^{\frac{(R_n^k)_{min}}{B_c}}$ and $A_n^k \geq 2$.
As noted from (\ref{eq-8}), $q^k$ plays a pivotal role to solve MSR metric.
$q^k$ is given by the waterfilling form,

\begin{equation}
  \begin{aligned}
    q^k &= \left[
      \frac{B_c}{\lambda} - \frac{A_2^k}{\Gamma_1^k}
      + \frac{A_2^k}{\Gamma_2^k} - \frac{1}{\Gamma_2^k}
    \right]_{\gamma^k}^{\infty}, \\
    \gamma^k &= \frac{A_2^k (A_1^k - 1)}{\Gamma_1^k} + \frac{A_2^k - 1}{\Gamma_2^k}.
  \end{aligned}
\end{equation}

\noindent To derive power budget $p^k$, Lagrangian multiplier method and bisection method are used,
and its upper bound is set to infinity and lower bound is set to $\gamma^k$, meaning that if the derived power budget $q^k$ is smaller than $\gamma^k$, $q^k$ is set to $\gamma^k$, otherwise keep $q^k$.

The optimal power allocation problem is solved with the mathematical solution (\textit{i.e.}, JRA), channel allocation problem remains to be solved.
The optimal channel allocation method can be driven by using exhaustive search (ES) method, however, it consumes extraordinarily long time to find the optimal allocation.
Previous works focused on solving an optimal power allocation problem, while leaving the channel allocation problem left with randomization, resulting an inefficient channel allocation.
On the other hand, we are integrating JRA, achieving optimal power assignment and applying our own DRL method to find a sub-optimal solution for channel allocation with more efficient way.

\section{Reinforcement Learning Algorithm}
\label{sec:rl}
In this paper, we utilized reinforcement learning algorithm to solve channel assignment problem by using multiple neural networks.
In this section, details of how channels are assigned to users, using fully-connected neural network (FCNN), convolutional neural network (CNN), and attention-based neural network (ANN) will be explained.

Exhaustive Search (ES) method is significantly more inefficient compared to the approach we propose.
ES method explores all possible channel allocations, represented by $\prod_{i=0}^{\frac{N-2}{2}}{C(N - 2i, 2)}$ where $N$ users are allocated and $N_k=2$, in a given environment and calculates the sum rate for each case.
However, as the number of users increases, the number of possible channel allocations grows exponentially.
Therefore, in realistic scenarios with many users, finding the maximum sum rate through optimal channel allocation by using ES method is highly inefficient and nearly impossible.
On the other hand, our proposed channel allocation method using DRL with replay memory utilizes previously learned experiences to train the model for near-optimal channel allocations demonstrating results close to the maximum sum rate, as shown in the Section~\ref{sec:eval}.
After training, our method consumes near-linear time complexity, typically within a second, compared to the ES method.

The fundamental of DRL formulation is to define state, action, and reward.
Each component is represented as $s_t$, $a_t$, and $r_t$ at time step $t$, corresponding to the state, action and reward, respectively.

A state is defined as a pair of user and channel information.
The state space is $N \times K \times F$, where $N$ is the number of users, $K$ is the number of channels, and $F$ is the number of features.
By forming the state with the space of $N \times K$, every possible combination of user and channel information can be represented.
The feature of the state represents the user and channel information itself and the number of features $F$ vary from one to three;
a state with $F=1$ contains CNR information, a state with $F=2$ contains CNR and distance information between users and the BS, and a state with $F=3$ contains CNR, distance, and channel assignment status information.
A CNR value of channel $k$ is represented as $CNR_k$, a distance between user $n$ and the BS is represented as $d_n$, and channel $k$'s assignment status is represented as $C_k$.
The value of the channel assignment status $C_k$ is equal to the number of users assigned to the following channel.
For instance, if the channel $k$ has zero user assignment, then the channel status is $C_k=0$.
When the user $n$ is assigned to channel $k$, then the status changes to $C_k=1$.
The channel status information allows the agent to be aware of assignable channels.
It is important to know which channel is assignable, due to each channel can hold fixed number of users.
In this case, since the $N_k$ is set to $2$, $C_k$ ranges from $0$ to $2$.

Under NOMA system, to solve the channel assignment problem, action is assigning a user to a channel. Therefore, the action can be represented with one user and one channel as $a_t = (n, k)$.
The selected state can also be interpreted as an action taken.

Reward is defined as a data rate (throughput) of each user.
The reward for channel $k$ at time step $t$ can be expressed in two cases as follow,

\begin{equation}
  r^k_t = 
  \begin{cases}
  R^k_1(s_t), & \text{if the user in $s_t$ is } \\
              & \text{first assigned to channel $k$}, \\
    R^k_2(s_t), & \text{otherwise},
  \end{cases}
\end{equation}

\noindent due to the constraint $N_k=2$.
The objective of the downlink NOMA system is to maximize the sum of all users' data rates (rewards) as follows,

\begin{equation}
  G_N^{MSR} = \max{\sum_{i=1}^{N}{r_i}}.
\end{equation}

To solve the channel assignment problem, let's say the set of actions taken is $\zeta = \{ a_1, a_2, \cdots, a_N \}$.
Given some state set $S$, the conditional probability of $\zeta$ is as follows,

\begin{equation}
  p_{\theta}(\zeta | S) = \prod_{i=1}^{N}{p_\theta(a_t | S)}
\end{equation}

\noindent where $\theta$ is a policy parameter, used to update the policy using loss function.
Variation of the reinforcement estimator~\cite{RL-bl} is used for the loss function, stabilizing the training by using the baseline model.
The loss function is defined as the average rewards corresponding to the state set $\zeta$ as below,

\begin{equation}
  Loss(\zeta | S) = \mathbb{E}_\zeta \left[ G_N^{MSR}(\zeta) \right],
\end{equation}

\noindent and the parameter $\theta$ from policy $p(\cdot)$ is updated via policy gradient method, utilizing the difference between the loss of online model and baseline model:

\begin{equation}
  \begin{aligned}
    &\nabla Loss(\zeta | S) \\
    &= \mathbb{E}_{\zeta} \left[ \left( Loss(\zeta | S) - Loss(\zeta^{bl} | S) \right)
    \nabla \log_{p_\theta}(\zeta | S) \right].
  \end{aligned}
  \label{eq-14}
\end{equation}

The algorithm used for training is as Algorithm.~\ref{alg}.
The training is performed on an episode basis, and performed until it reaches stopping criteria (line 1).
Each episode creates a user with randomized location (line 3).
Based on the coordinate of the corresponding user, the algorithm calculates the corresponding user's CNR.

After initializing the user information, the channel allocation for every user is excuted.
Every time step, the online model samples a user to allocate to a channel based on the model's probability distribution (line 6).
On the other hand, baseline model selects an user-channel pair with a highest probability (line 7).
Because selected user and channel can not be selected once again, the process of masking selected pairs is required, leading a more efficient calculation.
After repeating the mentioned steps, every user is allocated to all channels, then by using JRA method, powers are assigned, leading a sum rate.

When the actions are all taken, the result is saved into a replay memory $(\zeta, R, R^{bl})$ (line 9).
Sum rate of the online model $R$ and the baseline model $R^{bl}$ is calculated using the state sets $\zeta$ and $\zeta^{bl}$, by performing the JRA method.
Then, batch sized experiences are randomly sampled from replay memory (line 10), represented as $\delta$ and is as,

\begin{equation}
  \delta \leftarrow \{ \zeta_1, \zeta_2, \ldots, \zeta_{batch} \}.
  \label{eq-15}
\end{equation}

\noindent $\delta$ is used for calculating loss and the loss is derived using the gradient from (\ref{eq-14}) (line 11).

The used optimizer in this paper for updating the policy gradient is the Adam optimizer \cite{adam}.
With the optimizer, online model updates its own parameter $\theta$ as below,

\begin{equation}
  \theta \leftarrow Adam(\theta, \nabla Loss(S | \delta)).
  \label{eq-16}
\end{equation}

\noindent The update of the baseline model's parameter $\theta^{bl}$ is performed when the loss of the online model exceeds the loss of the baseline model as $\theta^{bl} \leftarrow \theta$
(\textit{i.e.}, online model's sum of rewards exceeds baseline model's sum of rewards) (line 13, 14).

By following the steps of (\ref{eq-15}) and (\ref{eq-16}), it is indeed true that the computational complexity increases,
due to the use of replay memory despite the problem being an on-policy problem.
The inefficiency of incorporating replay memory into an on-policy method arises from the nature of on-policy learnig.
On-policy methods involve iterative evaluation and updating of the policy based on the most recent experiences.
However, the experience replay memory contains past experiences that may not be representative of the current policy.
This mismatch means that experiences stored in the memory are more likely to be irrelevant to the updated current policy.
Still, the reason behind we incorporated the experience replay memory with REINFORCE algorithm is that we have noticed that utilizing the REINFORCE algorithm results in a narrow learning, unable to respond to generalized data.
On the other hand, the incorporation improves the model to be able to optimally allocate joint resources within various generalized data, although consumes more time on learning the policy.

After each episode is completed, validation is performed.
A validation set is created at the beginning of the training, when the NOMA environment is created.
The number of validation performed is equal to the number of validation seeds.
When the validation set is decided, maximum sum rate $R_{max}$ and minimum sum rate $R_{min}$ of each validation seed is calculated by using exhaustive search.
The deriven sum rates are used to calculate the error rate.
In order to calculate the error rate for each seed number, a sum rate from the baseline model $R^{bl}$ is used.
The error rate from the baseline model for validation $seed$ is defined as (line 18),

\begin{equation}
  Error_{seed}(R^{bl}) = \frac
  {R_{max} - R^{bl}} {R_{max} - R_{min}}.
  \label{eq-17}
\end{equation}

\noindent If the error rate is below the predefined threshold, the validation is considered passed (line 23, 24).
When every validation is passed for all seeds, the validation set is passed.

The stopping criteria for the training process are determined by whether the validation set passes and the loss threshold.
At the end of every episode, stopping criteria are evaluated.
When the validation set is passed and the loss between the online and baseline model is below the threshold, the stopping criteria are met.
Then the parameters from baseline model are saved and the training process terminates.

\begin{algorithm}[H]
  \caption{Training algorithm for channel assignment}
  \begin{algorithmic}[1]
    \Require State $S$, two models of NN; $p_\theta$ and $p_{\theta^{bl}}$, NOMA environment
    \Ensure Trained model NN; channel assignment

    \While{stopping criteria not met}
      \For{each episode}
        \State generate $N$ user profiles with random seeds

        \For{each step} \Comment{env.step(action)}
          \State $p_\theta(s_t | S) \gets$ output of NN
          \State $\zeta \gets $ sampling per $p_\theta(s_t | S)$
          \State $\zeta^{bl} \gets $ argmax sampling per $p_{\theta^{bl}}(s_t | S)$
        \EndFor

        \State replay memory $\gets (\zeta, R, R^{bl})$
        \State $\delta \gets \{ \zeta_1, \zeta_2, ..., \zeta_{batch} \}$
        \State $
        \begin{aligned}
          &\nabla Loss(\delta | S) \\
          &= \mathbb{E}_{\zeta}[(Loss(\zeta | S) - Loss(\zeta^{bl} | S)) \nabla \log{p_{\theta}(\zeta | S)}]
        \end{aligned}
        $
        \State $\theta \gets Adam(\theta, \nabla Loss(S | \delta))$ 

        \If{$Loss(\zeta | S) > Loss(\zeta^{bl} | S)$}
          \State $\theta^{bl} \gets \theta$
        \EndIf
      \EndFor

      \If{validation time step}
        \State run validations $Error_{seed}(R^{bl}) \gets \frac{R_{max} - R^{bl}}{R_{max} - R_{min}}$
        \If{$Error_{seed} \leq threshold$}
          \State validation passed
        \EndIf
      \EndIf

      \If{validation passed and $Loss < threshold$}
        \State save model $p_{\theta^{bl}}$
        \State break
      \EndIf
    \EndWhile
  \end{algorithmic}
  \label{alg}
\end{algorithm}

\section{Evaluations}
\label{sec:eval}
The simulations were conducted in the following simulation settings.
We assumed that there is a single BS which is an agent allocating and assigning channels and powers.
Around the BS, there are $N$ users scattered randomly from 50$m$ to 300$m$ which their minimum data rate is set to $(R_n^k)_{min} = 2 bps \slash Hz$ $\forall k = 1, \ldots, K$, $n = 1, \ldots, N$.
Since $N_k=2$, the number of channels is $K = N \slash 2$. 
The total bandwidth for the agent BS can use is $B_{tot} = 5MHz$.

The noise power spectral density for the environment is $N_0 = -170dBm \slash Hz$.
The channel response of $n$-th user assigned to $k$-th channel is represented as $h_n^k$ and it is defined as follows,

\begin{equation}
  h_n^k = g^k_n d^{-\alpha}_n
\end{equation}

\noindent where $g_n^k$ is a Rayleigh fading distribution corresponds to user $n$ and channel $k$, and $d^{-\alpha}_n$ is a distance loss between user $n$ and BS.
Here, $\alpha$ is a path loss coefficient which is set to $\alpha=2$.

The validation set was predefined before the training so that the maximum sum rate and minimum rate can be searched beforehand by exhaustive search.
Every training episode utilized an unique instance by using a seed number enabling the model to learn more generalized knowledge about the NOMA policy.
The validation was performed every 200 episodes of training to determine whether the loss and error rate from (\ref{eq-14}) and (\ref{eq-17}) met the stopping criteria.

In this section, to evaluate the performance of the proposed JRA-DRL method, various simulations
---effects of changes in learning rate, batch size, number of features, models and comparisons between JRA-DRL, JRA, and exhaustive search---
are analyzed.
Experimental parameters for evaluation of the framework were set as follows:
learning rates vary by 0.001, 0.0005, and 0.0001,
and batch sizes are 20, 40, and 80.
The simulations are conducted observing the actual sum rates and convergence speeds to assess the performance of the proposed framework.

\begin{figure}
  \centering
  \includegraphics[width=0.7\linewidth]{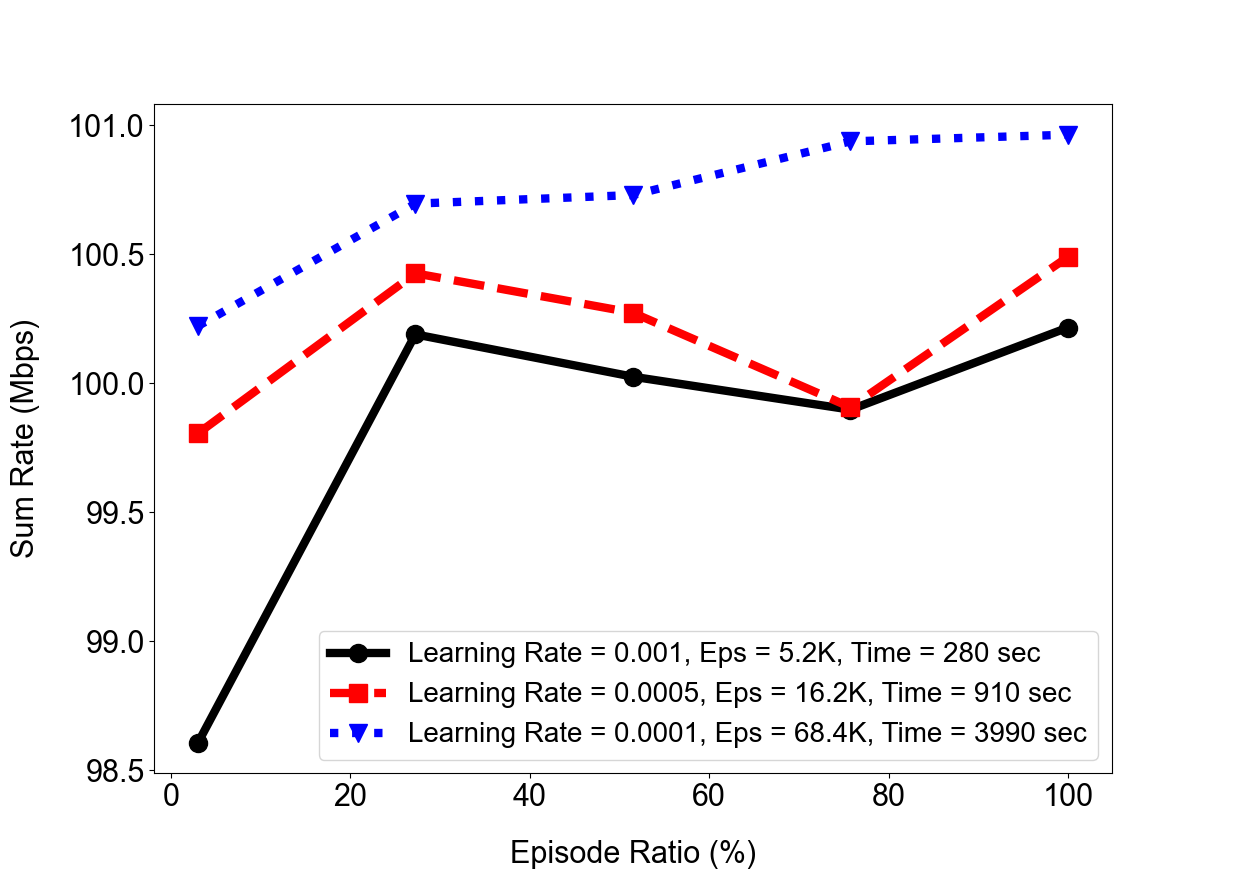}
  \caption{Sum rate comparison of different learning rates with $N \times K \times F = 6 \times 3 \times 3$, batch size = $40$ model = FCNN and $P_{T} = 12 W$.}
  \label{fig:2}
\end{figure}
Figure.~\ref{fig:2} illustrates achievable sum rates in different learning rates; 0.001, 0.0005, 0.0001.
The largest learning rate (0.001) converged the fastest, completing training in just 280 seconds; however, resulted in the lowest overall sum rate.
On the other hand, the smallest learning rate (0.0001) took the longest to converge, requiring 3,990 seconds to meet the stopping criteria, leading to the highest sum rate.
The learning rate of 0.0005 would be a great alternative to achieve the reasonable results, as it significantly reduces the time took for the training compared to the learning rate of 0.0001, while still achieving high sum rates.
However, the variance of the sum rates is large, therefore the fluctuations of sum rates were resulted.

\begin{figure}
  \centering
  \includegraphics[width=0.7\linewidth]{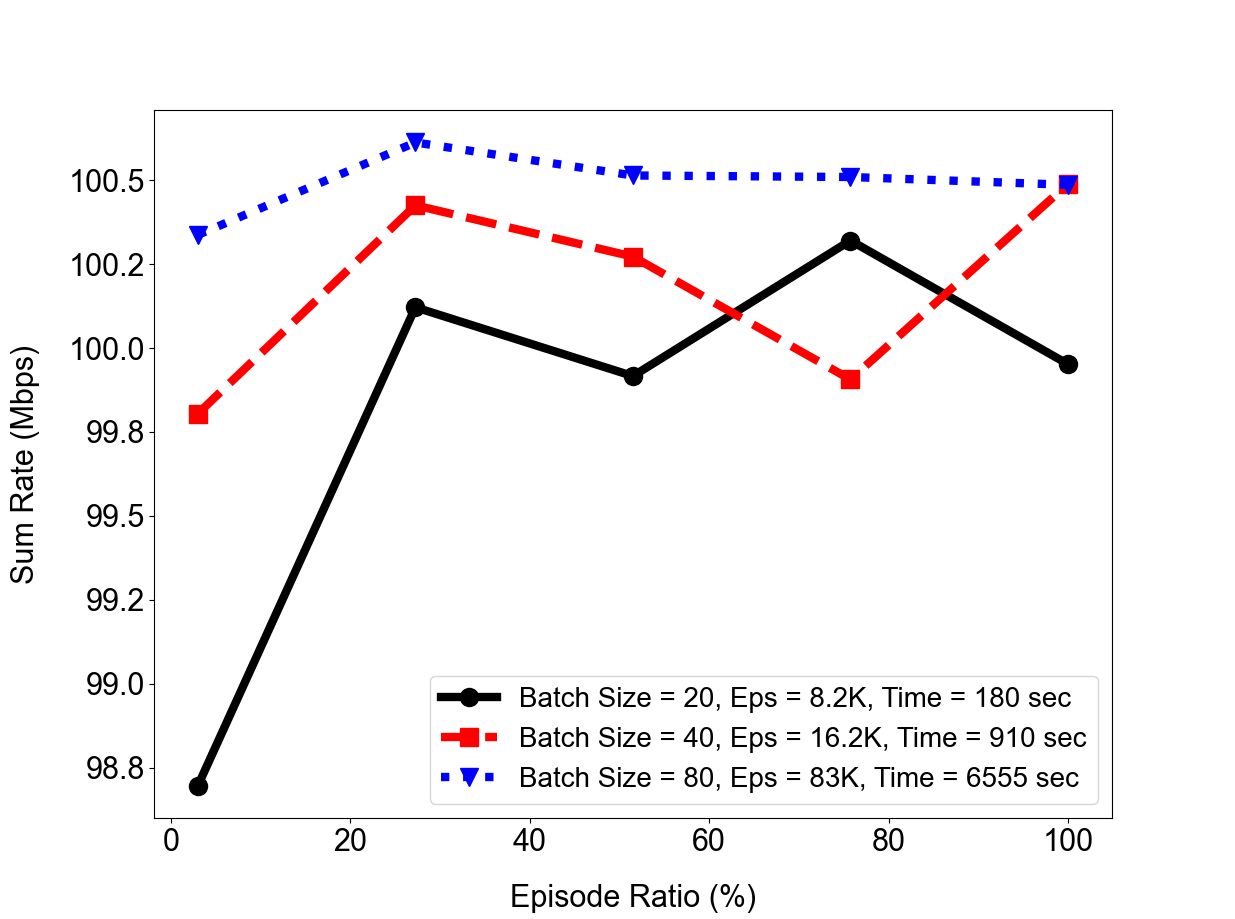}
  \caption{Sum rate comparison of different batch sizes with $N \times K \times F = 6 \times 3 \times 3$, learning rate = $0.005$, model = FCNN and $P_{T} = 12 W$.}
  \label{fig:3}
\end{figure}

Figure.~\ref{fig:3} shows the sum rates in different batch sizes; 20, 40, and 80.
As the comparison of different learning rates from Figure.~\ref{fig:2} did, the results for different batch sizes were very similiar.
The smallest batch size of 20 was able to finish the training the fastest, nevertheless it yielded the lowest overall performance.
Compared to this, the sum rate of the largest batch size of 80 was the highest, taking the longest time to complete the training.
Similarly with the comparison of different learning rates, the intermediate batch size of 40 completed the training relatively quickly compared to the larger batch size of 80, while yielding high sum rates.

\begin{figure}
  \centering
  \includegraphics[width=0.7\linewidth]{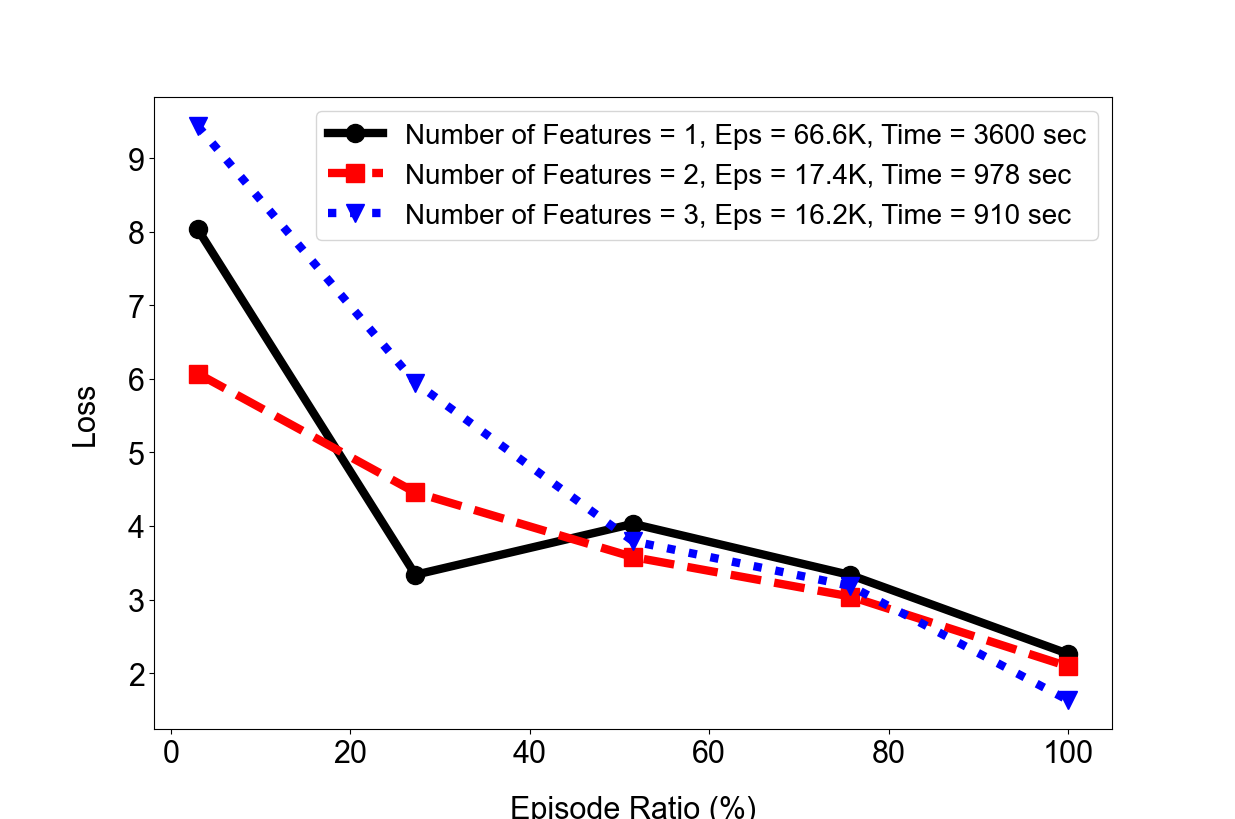}
  \caption{Loss comparison of different number features with $N \times K = 6 \times 3$, learning rate = $0.005$, batch size = $40$, model = FCNN and $P_{T} = 12 W$.}
  \label{fig:4}
\end{figure}

Figure.~\ref{fig:4} represents the comparison of loss convergences when using different number of features for the state;
($CNR_k$), ($CNR_k, d_n$), and ($CNR_k, d_k, C_k$) for all $k=1, \ldots, K$ and $n=1, \ldots, N$.
At the beginning of the training, all three agents with different state spaces have high loss values.
Then as the number of training episodes increases, the loss converges towards zero.
The agent with three features requires the most training time to converge, meaning that it took the longest to figure out the meaning of the state.
In contrast, the agents with one feature and two features require much less training time to converge.
Nevertheless, in the middle of the training, the agent with one feature's loss value abruptly increases, indicating unstable learning.

\begin{figure}
  \centering
  \includegraphics[width=0.7\linewidth]{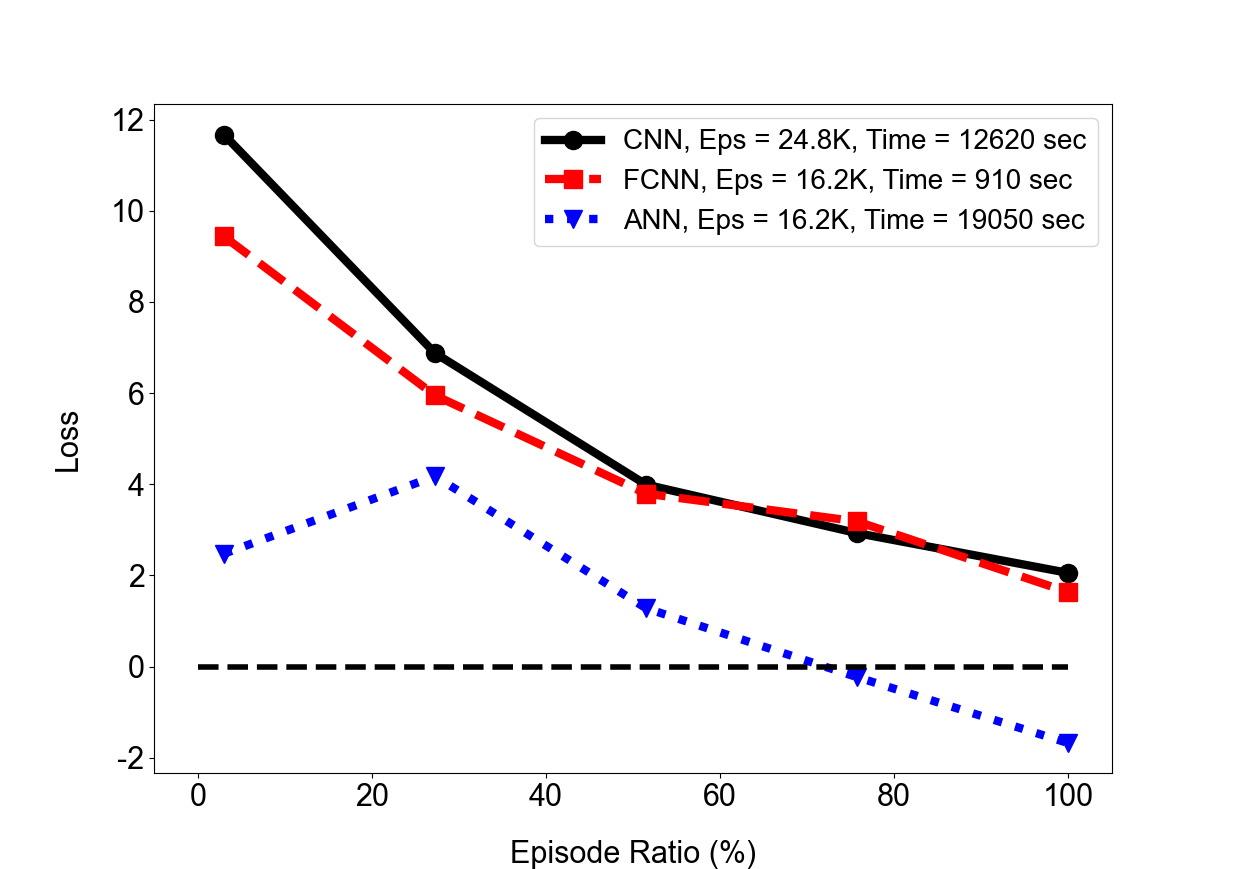}
  \caption{Loss comparison of different models with $N \times K \times F = 6 \times 3 \times 3$, learning rate = $0.005$, batch size = $40$ and $P_{T} = 12 W$.}
  \label{fig:5}
\end{figure}

Figure.~\ref{fig:5} represents the comparison of loss convergences when using different models; FCNN, ANN, and CNN.
The loss values for the agents using the FCNN and CNN models decrease gradually and converge towards zero, while the CNN model requires the longest training episodes.
However, although the ANN model required the fewest episodes to converge among the three models, its loss oscillates constantly.
This means that learning the NOMA policy from the baseline model is not sufficiently stable, leading to consistent changes in the sign of the loss values.
The reason for implementing a baseline model to policy gradient methods is to stabilize training and reduces the variance of the gradient estimates.
Since the baseline model provides a stable reference value, monitoring the sign of the loss value enables the agent to stabilize the learning and reduce the variance.
However, for the ANN model, frequent changes in the baseline model resulted in unstable training and learning.
Furthermore, despite the fact that the number of episodes required for training the ANN and the FCNN models is nearly identical, the actual time consumed by the ANN model significantly exceeded the time when using the FCNN model.

\begin{figure}
  \centering
  \includegraphics[width=0.7\linewidth]{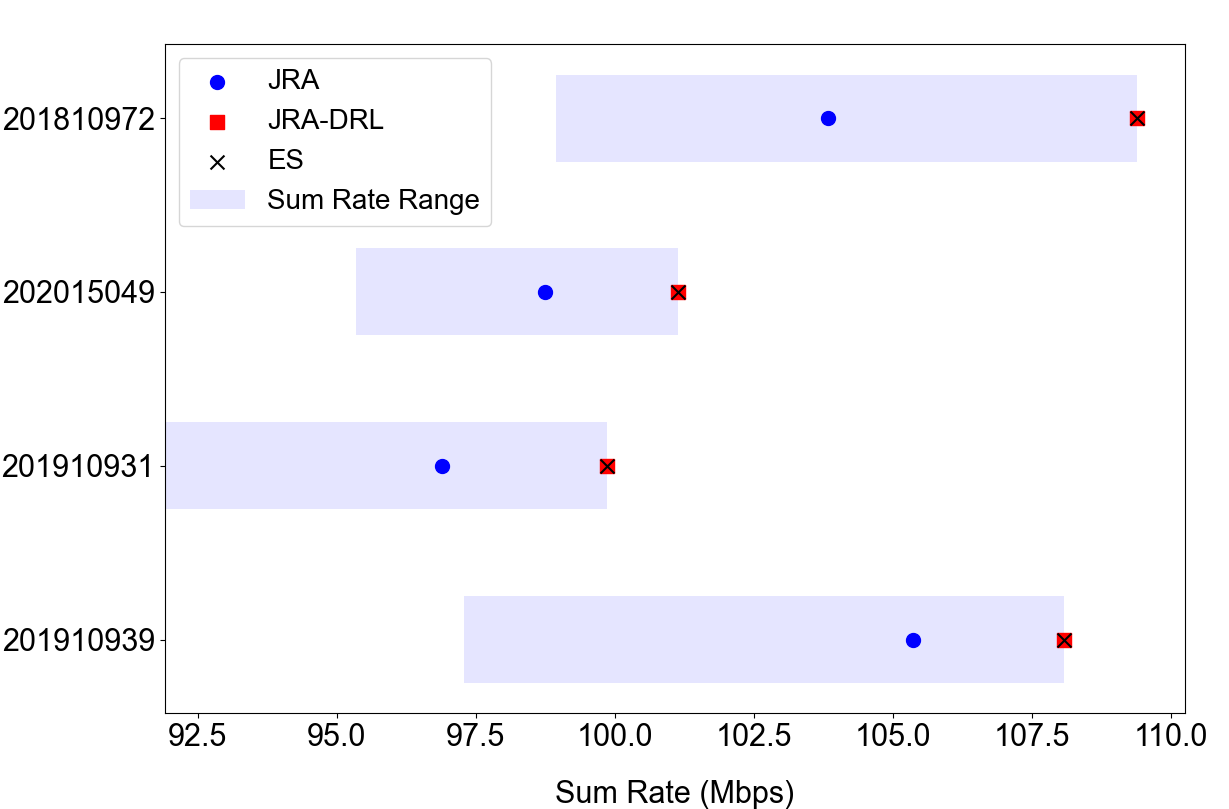}
  \caption{Sum rate performance comparison of different training algorithms.}
  \label{fig:6}
\end{figure}

Figure.~\ref{fig:6} illustrates the performance comparison between three methods: the exhaustive search (ES) method, the JRA method, and the proposed JRA-DRL method.
The methods are evaluated using the trained model and the evaluation is based on the resulting sum rates in four different NOMA environments (seeds).
The four different NOMA environments used for the validation (or evaluation) are entirely different with the environments used for the training.
The key difference is that the validation set comprised more generalized environmental settings, which makes it challenging for the standard policy gradient method to adapt effectively.
Obviously, the exhaustive search method achieves the maximum attainable sum rates across all seed numbers.
The JRA method achieves high sum rates, though not the highest.
The proposed method, JRA-DRL method, achieves sum rates that are very close to the maximum attainable sum rates in all four simulations.
This shows that implementing experience replay into the policy gradient method stimulates the agent to adapt effectively to generalized environment.

\begin{figure}
  \centering
  \includegraphics[width=0.7\linewidth]{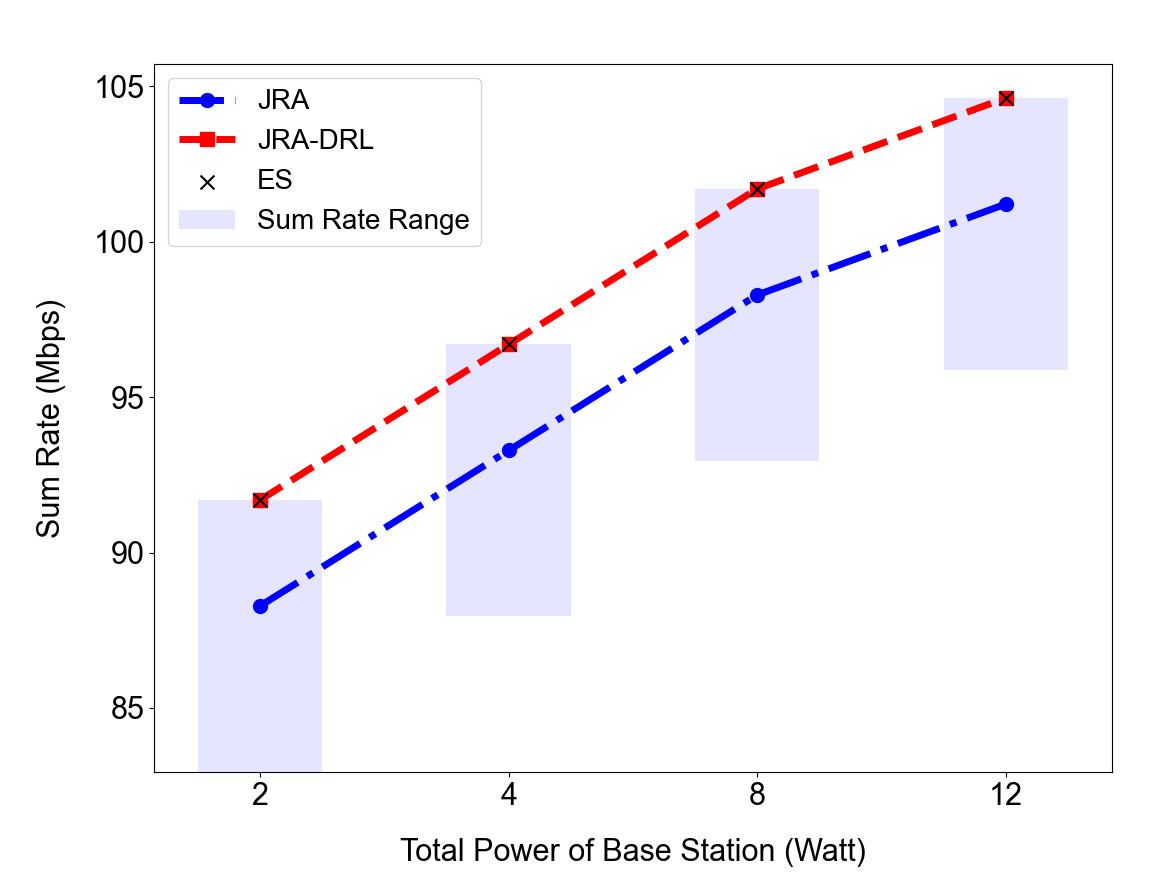}
  \caption{Sum rate comparison of different training algorithms with different total power for base station.}
  \label{fig:7}
\end{figure}

Figure.~\ref{fig:7} illustrates the sum rate comparison of different training algorithm with different total power $P_{T}$ for BS; $2 W$, $4 W$, $8 W$ and $12 W$.
The above shows that as the total power for BS $P_{T}$ increases, the attainable sum rate also increases.
As the JRA-DRL method yields higher sum rates near to maximum attainable sum rates from exhaustive search than the JRA method,
the JRA-DRL method is proven to be able to achieve superior performance in all power levels.

\begin{figure}
  \centering
  \includegraphics[width=0.7\linewidth]{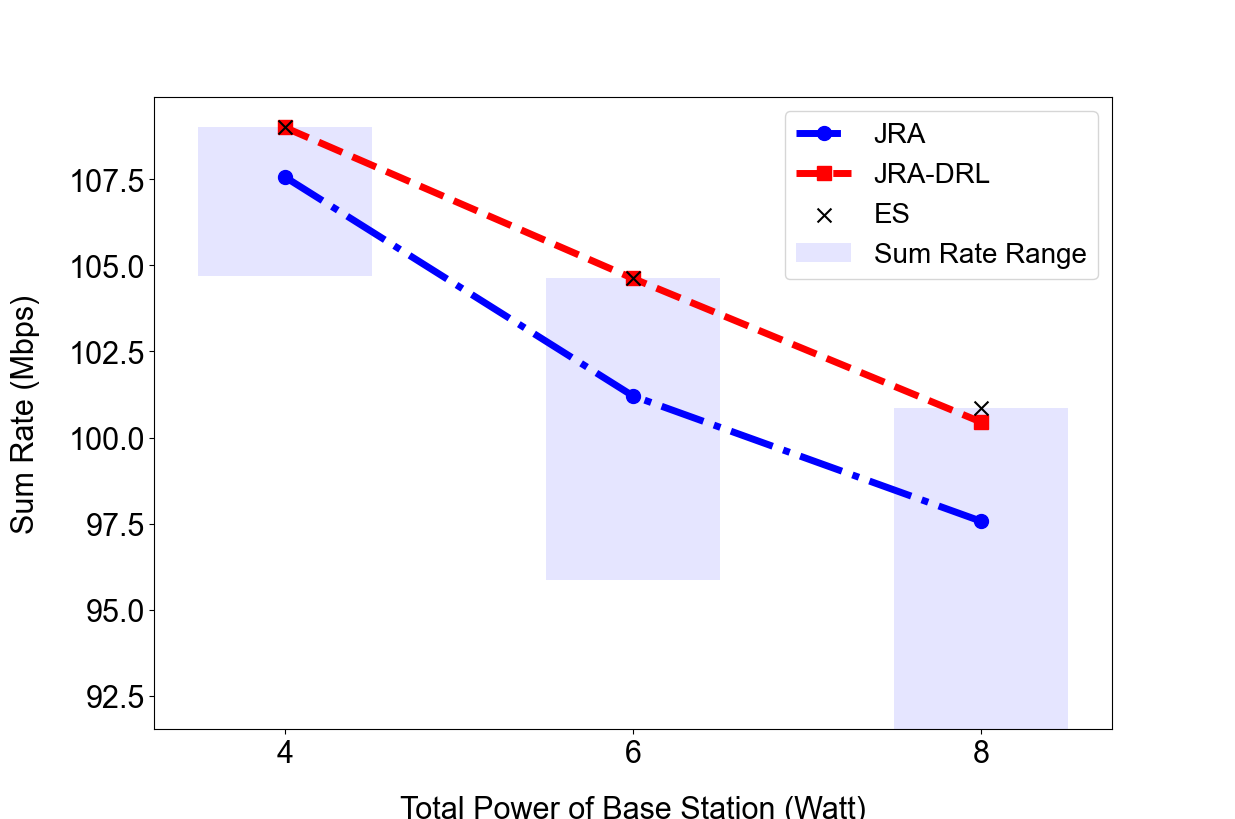}
  \caption{Sum rate comparison of different training algorithms with different number of NOMA users.}
  \label{fig:8}
\end{figure}

Figure.~\ref{fig:8} illustrates the sum rate comparison of different training algorithms with different number of NOMA users $N$; 4, 6 and 8.
As the number of the NOMA users increases, the attainable sum rates decrease, because the users are forced to shared limited joint resources, resulting in a lower sum rates.
In all simulation results, the JRA-DRL method achieves higher sum rates than the JRA method, which are very close to the sum rates of exhaustive search method.

\begin{figure}
  \centering
  \includegraphics[width=0.7\linewidth]{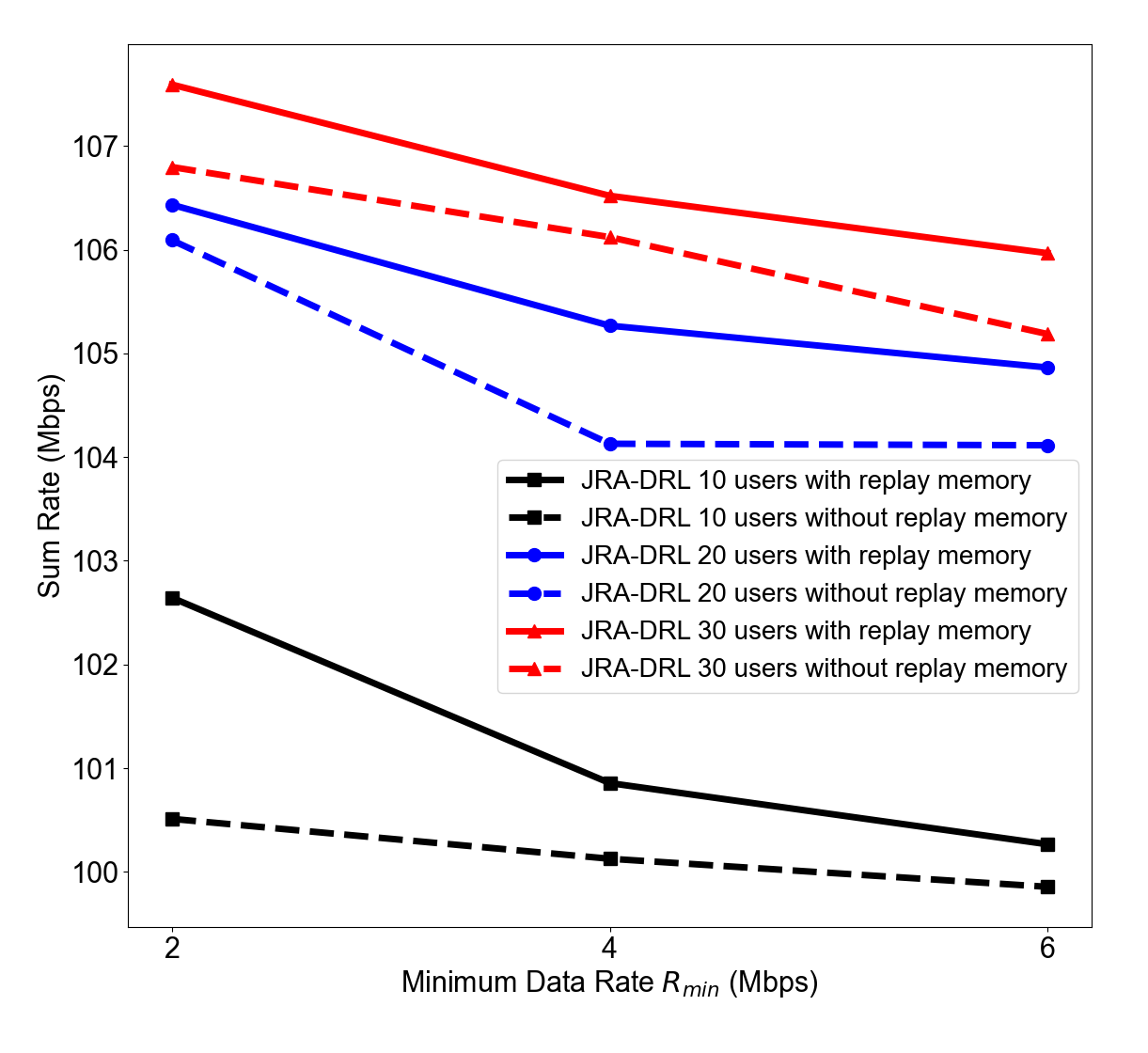}
  \caption{Sum rate comparison for JRA-DRL method with and without replay memory of different state sizes when requiring different minimum data rate ($R_{min}$).}
  \label{fig:9}
\end{figure}

Additionally, the sum rate performance of applying larger state sizes was evaluated with respect to minimum data rate $R_{min}$ for all users,
and the result is shown in Figure.~\ref{fig:9}. 
As mentioned in Section~\ref{sec:rl}, the number of possible channel allocations grows exponentially when the number of users and channels increases.
Due to this nature, the calculation of the error rate from Algorithm~\ref{alg} was omitted in this simulation.
As the minimum data rate $R_{min}$ increases, the sum rate gradually decreases for all different states sizes of $N \times K = 10 \times 5$, $N \times K = 20 \times 10$, and $N \times K = 30 \times 15$.
However, compared to Figure.~\ref{fig:8}, the simulation for smaller state sizes, increasing the state size leads to an improvement in sum rate performance.

Finally, Figure.~\ref{fig:9} also illustrates the performance comparison between JRA-DRL method with replay memory and without replay memory.
The proposed JRA-DRL method with replay memory shows the higher sum rate performance over the JRA-DRL method without replay memory for all state sizes and minimum data rates $R_{min}$.
This not only demonstrates that our proposed method can handle large input spaces, but also shows that the incorporation of the policy gradient method with replay memory ensures the generalization of policy learning, leading to higher sum rate performance.

As demonstrated by the simulations shown in this section, the proposed JRA-DRL method exhibits superb performance.
The JRA method is capable of allocating users with optimal powers but lacks the ability to assign users to adequate channels.
Due to this, the JRA method performs above average but is incapable to reach near to the highest performance.
In contrast, the proposed JRA-DRL method reaches near-maximum performance with reasonable training time, also showing the ability to adapt to completely new (or generalized) environments by using the experience replay.

\section{Conclusions}
In this paper, we propose a reinforcement learning-based framework to solve channel and power allocation problem in a downlink NOMA system,
and provide various simulation results.
The framework takes two steps which are channel allocation and power assignment.
At each time step, the model considers the current channel allocation status of users and decides which user should be allocated to which channel.
The channel allocation is performed by integration of replay memory and the REINFORCE algorithm which enables more generalized learning of the NOMA policy.
After the initial channel allocation step is completed, the subsequent power assignment step is carried out using the JRA method, which has been proven to be an effective solution for the power optimization problem.

The simulations were conducted with respect to the number of state features, batch sizes, types of models, and learning rates.
Overall, the simulations demonstrates that the proposed framework can successfully learn policy from NOMA system with fast convergence and has the ability to handle comprehensive data.

\bibliographystyle{plainnat}  
\bibliography{references}

%
%
%
%

\end{document}